%% file: main.tex
\documentclass[11pt]{article}

\usepackage{authblk}

\usepackage{amssymb}
\usepackage{amsmath}
\usepackage{upgreek}
\usepackage{stackengine}

\usepackage[LGR,T1]{fontenc}
\DeclareSymbolFont{ttgreek}{LGR}{cmtt}{m}{n}
\DeclareMathSymbol{\ttalpha}{\mathord}{ttgreek}{`a}
\DeclareMathSymbol{\ttmu}{\mathord}{ttgreek}{`m}
\DeclareMathSymbol{\tttau}{\mathord}{ttgreek}{`t}
\DeclareMathSymbol{\ttdelta}{\mathord}{ttgreek}{`d}
\DeclareMathSymbol{\ttDelta}{\mathord}{ttgreek}{`D}

\usepackage{caption}
\usepackage{subcaption}

\usepackage[breaklinks=true,colorlinks]{hyperref}
\usepackage{booktabs}
\usepackage{multirow}
\usepackage{float}

\usepackage[dvipsnames]{xcolor}
\definecolor{green2}{RGB}{0, 194, 0}

\usepackage{utfsym}
\newcommand{\xterm}{\scalebox{0.75}{\usym{2613}}}

\newcommand{\etal}{\emph{et al.}}
\newcommand{\ie}{\textit{i}.\textit{e}., }
\newcommand{\eg}{\textit{e}.\textit{g}., }

\newcommand{\first}[1]{{\color{blue} \textbf{#1}}}
\newcommand{\second}[1]{{\color{green2} #1}}
\newcommand{\third}[1]{{\color{red} #1}}

\input{pcr_defs}

\begin{document}

\title{On the representation and methodology for wide and short range head pose estimation}

\author[1, 3]{Alejandro Cobo}
\author[1, 3]{Roberto Valle}
\author[2, 3]{Jos\'e M. Buenaposada}
\author[1, 3]{Luis Baumela}

\affil[1]{Universidad Politécnica de Madrid,
            Campus de Montegancedo s/n, 
            Boadilla del Monte,
            28660, 
            Madrid,
            Spain}

\affil[2]{Universidad Rey Juan Carlos,
            Calle Tulipán s/n, 
            Móstoles,
            28933, 
            Madrid,
            Spain}

\affil[3]{\texttt{http://www.dia.fi.upm.es/$\sim$pcr}}

\date{}

\maketitle            
            
\begin{abstract}
Head pose estimation (HPE) is a problem of interest in computer vision to improve the performance of face processing tasks in semi-frontal or profile settings. Recent applications require the analysis of faces in the full $360^\circ$ rotation range. Traditional approaches to solve the semi-frontal and profile cases are not directly amenable for the full rotation case.
In this paper we analyze the methodology for short- and wide-range HPE and discuss which representations and metrics are adequate for each case. We show that the popular Euler angles representation is a good choice for short-range HPE, but not at extreme rotations. However, the Euler angles' gimbal lock problem prevents them from being used as a valid metric in any setting. 
We also revisit the current cross-data set evaluation methodology and note that the lack of alignment between the reference systems of the training and test data sets negatively biases the results of all articles in the literature.  We introduce a procedure to quantify this misalignment and a new methodology for cross-data set HPE that establishes new, more accurate, SOTA for the 300W-LP/Biwi benchmark.
We also propose a generalization of the geodesic angular distance metric that enables the construction of a loss that controls the contribution of each training sample to the optimization of the model. 
Finally, we introduce a wide range HPE benchmark based on the CMU Panoptic data set.
\end{abstract}


\section{Introduction}
\label{sec:introduction}

\begin{figure}
    \centering
    \includegraphics[width=0.75\textwidth]{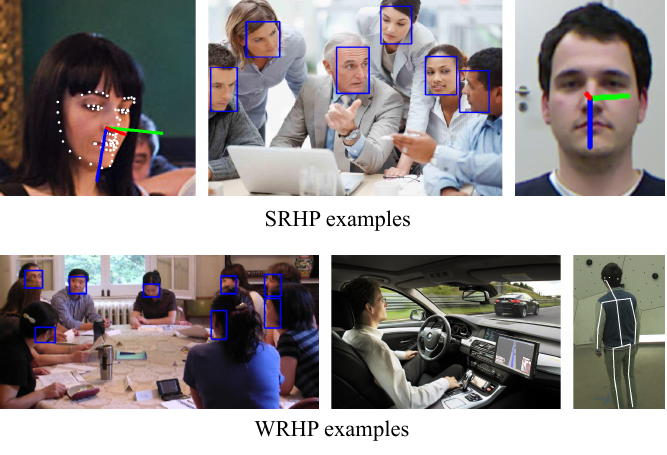}
    \caption{Applications involving SRHP and WRHP configurations. Images from 300W-LP~\cite{Zhu19b}, WIDER Face~\cite{Yang16b}, Biwi~\cite{Fanelli13} and CMU Panoptic~\cite{Joo17} data sets.}
    \label{fig:pose_examples}
\end{figure}

Head Pose Estimation (HPE) aims to compute the three-dimensional orientation of human heads in images or videos. This is a problem that has been widely studied in computer vision (CV) because it is a key element in many facial analysis workflows~\cite{Abate22}. Present state-of-the-art (SOTA) solutions achieve a Mean Absolute Euler angle Error (MAE) below $4^{\circ}$~\cite{Valle21,Albiero21,Martyniuk22} in realistic cross-data set experiments with Short Range Head Poses (SRHP) involving semi-frontal and profile faces with yaw angles roughly between $\pm90^{\circ}$. This is appropriate for boosting the performance of facial analysis algorithms that recognize faces~\cite{Barra18,Barra20}, estimate facial attributes~\cite{Ranjan19} or facial expressions~\cite{Valstar17} (see Fig.~\ref{fig:pose_examples}). However, they fail in the context of, \eg driver monitoring~\cite{Sumit17}, group interactions~\cite{Joo17} and surveillance~\cite{Rahmaniar22} applications, involving Wide Rage Head Poses (WRHP) with yaw angles of up to $\pm180^{\circ}$ (see Fig.~\ref{fig:pose_examples}). For this reason, the WRHP estimation problem is a topic of renewed interest~\cite{Beyer15,Zhou20}.

The immediate application of the traditional SRHP training and evaluation methodology to the WRHP setting produces HPE models with poor performance.
The most popular approach to SRHP estimation involves the use of Euler angles to represent the orientation of the head and the Mean Absolute Error (MAE) to measure the estimation error~\cite{Abate22}. Although Euler angles have been widely adopted in the robotics community, when used in a WRHP setting, their representation is discontinuous~\cite{Zhou19} and ambiguous, in presence of the so-called \emph{gimbal lock}. In this case, the MAE of two orientations is not a good measure of their distance because two nearby rotations may have a large MAE (see Sect.~\ref{sec:method_representation}). Further, a discontinuous orientation representation poses a difficult learning problem and in practice models with such representation perform worse than those using a continuous one~\cite{Zhou19}. So, to train  a WRHP estimation model, we need a continuous orientation representation and a proper metric~\cite{Huynh09} to define the training algorithm loss function and performance evaluation measure.

\begin{figure}
    \centering
    \includegraphics[width=0.9\textwidth]{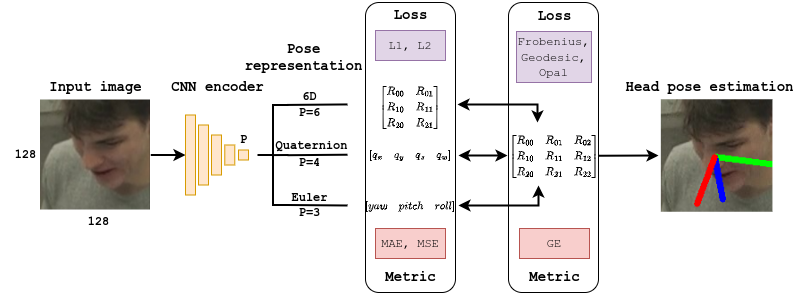}
    \caption{Concept diagram of our analysis. Given an RGB image  containing a cropped face, the HPE estimation algorithm produces a pose representation, $\vp$. Independently of the internal representation used by the model, predictions can be converted to Euler/quaternion angles or rotation matrices and measure the estimation error using different metrics.}
    \label{fig:model_architecture}
\end{figure}

In this paper we study the methodology for training and evaluating HPE algorithms to propose a suitable representation, loss function and evaluation metric for both SRHP and WRHP problems (see Fig.~\ref{fig:model_architecture}). The gimbal lock is the reason given for discarding the use of the Euler angles representation~\cite{Hsu19,Martyniuk22,Hempel22,Cao21} in HPE problems. In Sect.~\ref{sec:experiments_synthetic} and~\ref{sec:experiments_short_hpe} we analyze this issue and show that the gimbal lock is not a problem for HPE representation. In fact, Euler angles are an excellent choice in SRHP estimation problems. It is the issue of discontinuity arising at extreme rotations that prevents Euler angles, and the traditional quaternion-based alternative, from being used in the WRHP setting. 

Concerning the evaluation metric, we show that it is the gimbal lock issue that prevents MAE, the most popular error evaluation measure, \eg~\cite{Abate22,Hsu19,Zhou20,Hempel22}, 
from being used for HPE in SRHP and WRHP settings. We propose the use of the angular geodesic distance~\cite{Huynh09}, valid in any angular range. Based on the interpretability of this metric, we propose Opal, a loss function that allows to control the contribution of each sample to the model training as a function of its geodesic distance to the ground truth.

The most realistic methodology for evaluating HPE is a cross-data set procedure, in which the model is trained with one data set and evaluated with a different one~\cite{Valle21,Albiero21,Martyniuk22,Hempel22,Li23a}. In this context, a critical issue that has been ignored so far in the literature is the lack of alignment between training and test data set reference systems. In Sect.~\ref{sec:nmethod_alignment} we introduce a procedure to quantify this misalignment and establish a new SOTA in the popular cross-data set HPE evaluation methodology that uses 300W-LP~\cite{Zhu19b} for training and Biwi~\cite{Fanelli13} for testing.

In summary, our contributions are:
\begin{itemize}
\item We analyze the HPE pipeline in terms of the orientation representation (Euler angles, Quaternions or rotation matrix), and distance used, and discuss when each are amenable either for SRHP or WRHP.
\item We introduce a new loss function, Opal, based on a generalization of the geodesic distance between two rotation matrices, which allows us to weigh the contribution of each image in the data set to the minimization depending on the distance to the ground truth.
\item We propose a method to estimate the misalignment between train and test data set reference systems providing a more accurate evaluation protocol in cross-data set settings. We establish a new, more accurate, SOTA when training in 300W-LP and evaluating in Biwi.
\end{itemize}


\section{Related work}
\label{sec:related-work}

Object pose estimation is a topic of interest not only in the face processing area~\cite{Beyer15, Zhou20, Albiero21, Valle21} but in CV in general~\cite{Jiang22, Ponimatkin22}. When we talk about HPE, we refer to the estimation of the 3 DoF representing the orientation of the head w.r.t. the camera (see Fig.~\ref{fig:hpe_config}).
We can organize the HPE literature in two groups. The traditional SRHP approach, when the yaw angle is in the range $[-90^{\circ},90^{\circ}]$ (frontal to profile faces)~\cite{Ruiz18, Yang19, Hsu19, Albiero21, Valle21, Cao21, Hempel22, Martyniuk22} and the WRHP~\cite{Beyer15, Zhou20} when the range of the yaw angle is in $[-180^{\circ},180^{\circ}]$. 

To train a HPE model we need to define the orientation representation, the loss function and the evaluation metric between ground truth and estimated poses. In this section we review different approaches in the literature for each of these elements.

\subsection{Head pose representation}
The usual orientation representation in HPE is based on Euler angles~\cite{Abate22,Valle21,Zhou20,Ruiz18,Yang19}. The rotation matrix is broken down into three rotations around each camera axis, corresponding to the pitch, yaw and roll pose angles (see Fig.~\ref{fig:hpe_config}). In this configuration the gimbal lock problem (see Sect.~\ref{sec:method_representation}) occurs when the face is in profile orientation, so the yaw angle is $90^{\circ}$ or $-90^{\circ}$. 
Quaternions are also a popular alternative in object pose~\cite{Jiang22} and in HPE~\cite{Hsu19} because they have no gimbal lock.
The most significant problem in WRHP estimation is that both Euler angles and quaternion representations are discontinuous, which makes learning with deep networks difficult~\cite{Beyer15, Zhou19}. In fact, 
any 3D orientation representation with less than 5 parameters is discontinuous~\cite{Zhou19}. 

Different representations for HPE have been introduced in the literature with the aim of solving the discontinuity problem~\cite{Beyer15, Zhou19, Cao21}. 
Beyer~\etal~\cite{Beyer15} propose a continuous representation for the head yaw angle, termed \emph{biternion}, which is modeled with the vector $\vy=[cos(\phi), sin(\phi)]$, the first column of a 2D rotation matrix.
This was later generalized to a continuous 3D rotation model with a 6D representation composed of the first two columns of a 3D rotation matrix~\cite{Zhou19}, that has been used in the general object pose estimation~\cite{Ponimatkin22} as well as in SRHP~\cite{Hempel22, Martyniuk22}. We will denote this representation in our paper as 6D.
%
Cao~\etal~\cite{Cao21} propose a representation based on the full rotation matrix in the context of HPE. 

In Section~\ref{sec:experiments} we show that, although the 6D representation is immune to gimbal lock and discontinuities and therefore preferable in a WRHP setting, plain Euler angles are the best choice for SRHP estimation.

\subsection{Loss functions and metrics for HPE}

3D rotations arise in many contexts in the scientific literature and there are different functions to measure the distance between two rotations~\cite{Huynh09}. We need them to define the loss functions used to train the models and to evaluate their performance.

There are a few metrics to compare rotations in terms of Euler angles representation. Let the vector $\vp=(\alpha_p,\alpha_y,\alpha_r)$ represent a pose configuration with the three Euler angles and $\hat\vp$ the pose estimated by a model. The MAE and Mean Squared Error (MSE) of an estimation are given by the summation extended to the data set of respectively $g_{MAE}(\hat\vp,\vp)=||\hat\vp-\vp||_1$ and $g_{MSE}(\hat\vp,\vp)=||\hat\vp-\vp||_2^2$, where $||\cdot||_p$ denotes the $L_p$ norm. There are other metrics that take into account the periodicity of angular representation. In this case the i-normed difference between two angles $a$, $b$, is given by $d_i(a,b)=\min\{||a-b||_i,||360^\circ-|a-b|\, ||_i\}$, and define alternative metrics such as the Euclidean Distance between Euler angles~\cite{Huynh09}, $g_{EUC}(\hat\vp,\vp)=||\vd_1(\hat\vp,\vp)||_1$, and the, so-called, wrapped yaw distance~\cite{Zhou20} $g_w(\hat\alpha_y,\alpha_y)=d_2(\hat\alpha_y,\alpha_y)$.

As we discuss in Sect.~\ref{sec:method_representation}, all distances based on Euler angles can lead to erroneous results in presence of gimbal lock. In the usual pitch-yaw-roll representation used in HPE, this happens when the face is close to a profile configuration, $\alpha_y\approx\pm90^\circ$.

Other metrics directly compare rotation matrices and have also been used in HPE~\cite{Martyniuk22, Cao21}.
Some are based on the Frobenius norm, such as the \emph{chordal distance}~\cite{Hartley13}, $g_F(\mR,\hat\mR)=||\mR-\hat\mR||_F$,  its squared form~\cite{Cao21},  the deviation from identity matrix, $g_{DI}(\hat{\mR}, \mR)= ||\mI - \hat{\mR}\mR^T||_F$~\cite{Huynh09,Martyniuk22}, or the \emph{geodesic distance}~\cite{Huynh09,Zhou19,Hartley13}, 
\begin{equation}
     g_{GE}(\hat{\mR}, \mR)=\cos^{-1}\left(\frac{tr(\hat{\mR}\mR^T) - 1}{2}\right).
\label{eq:geodesic}
\end{equation}

Both $g_{GE}$ and $g_{DI}$ are proper metrics in SO(3) and quantify the rotation required to bring $\hat{\mR}$ in coincidence with $\mR$~\cite{Huynh09}. The former, however, has a direct interpretation as an angle. Thus, we will use the geodesic distance between the predicted rotation matrix for the $i$-th image, $\hat{\mR}_i$, and its ground truth, $\mR_i$, for all samples in a data set to evaluate the quality of a model. We denote this measure as \emph{Geodesic Error} (GE):
\begin{align}
f_{GE} = \frac{1}{N} \sum_{i=1}^N g_{GE}(\hat{\mR}_i, \mR_i),
    \label{eq:geodesic_error}
\end{align}
where $N$ is the number of images in the data set.

\begin{table}
    \centering
    \footnotesize
    \begin{tabular}{l|c|c|c}
        \toprule
        \multirow{2}{*}{Method} & Continuous & Gimbal lock-free & Wide range \\
        & representation & metric & evaluated \\
        \midrule
        HopeNet~\cite{Ruiz18} & \color{red} \xterm & \color{red} \xterm & \color{red} \xterm \\
        FSA-Net~\cite{Yang19} & \color{red} \xterm & \color{red} \xterm &  \color{red}\xterm \\
        WHENet~\cite{Zhou20} & \color{red} \xterm & \color{red} \xterm & \color{green} \checkmark \\
        TriNet~\cite{Cao21} & \color{green} \checkmark & \color{green} \checkmark & \color{red} \xterm \\
        6DRepNet~\cite{Hempel22} & \color{green} \checkmark & \color{red} \xterm & \color{red} \xterm \\
        \midrule
        \textbf{Ours} & \color{green} \checkmark & \color{green} \checkmark & \color{green} \checkmark \\
        \bottomrule
    \end{tabular}
    \caption{Comparison of the most relevant HPE works with our methodology.}
    \label{tab:sota_summary}
\end{table}

In summary, to build a model adequate for WRHP estimation we have to select a representation that is continuous, use a correct metric and evaluate it in a wide range data set. To the best of our knowledge, as we can see in Table \ref{tab:sota_summary}, we are the first to use a continuous representation with an accurate metric free from gimbal lock, and evaluate results using a WRHP data set.


\section{Representation and methodology for HPE}
\label{sec:method}

The accuracy of a HPE algorithm depends on the range of possible head orientations and the right choice of rotation representation, error metric and evaluation methodology. In this section we first revisit the gimbal lock problem and discuss its impact in the Euler angle representation and the MAE. 
Finally, we discuss and solve the problem of miss-alignment in cross-data set HPE evaluation.

\subsection{The gimbal lock}
\label{sec:method_representation}

\begin{figure}
    \centering
    \tiny
    \subfloat[Example configuration of head and camera reference systems]{
    \label{fig:hpe_config:a}
    \includegraphics[width=0.6\textwidth]{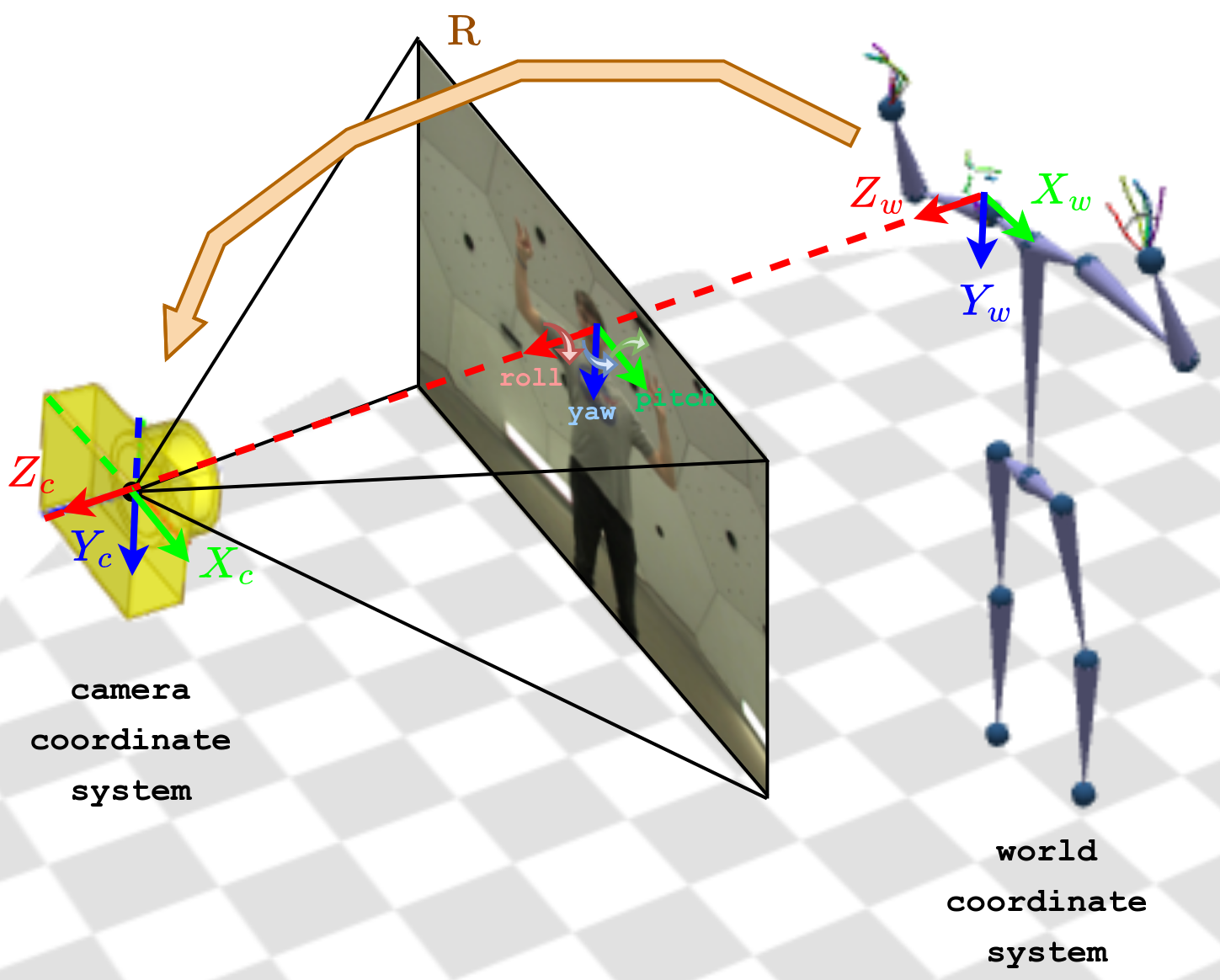}}
    \subfloat[WRHP examples]{
    \label{fig:hpe_config:b}
    \begin{tabular}[h]{c}
    \stackunder[1.5pt]{\includegraphics[width=0.09\textwidth]{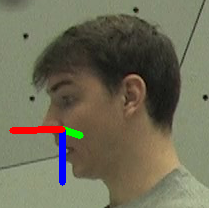}}{[-68.93, -18.72, -16.92]}\\
    \stackunder[1.5pt]{\includegraphics[width=0.09\textwidth]{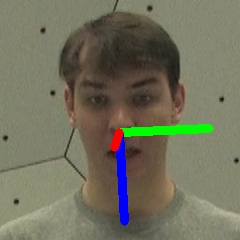}}{[-3.98, -9.73, -4.39]}\\
    \stackunder[1.5pt]{\includegraphics[width=0.09\textwidth]{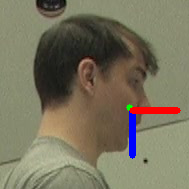}}{[96.38, 43.81, 42.48]}\\
    \stackunder[1.5pt]{\includegraphics[width=0.09\textwidth]{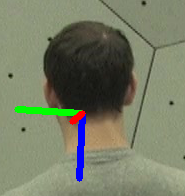}}{[-171.70, -7.68, 2.10]}\\
    \end{tabular}}
    \caption{WRHP means estimating the rotation matrix $\mR$ to align the reference frame of the head with that of the camera. We show some WRHP results projecting a 3D axis onto the image plane coordinates. The text below represent the [yaw, pitch, roll] angles.}
    \label{fig:hpe_config}
\end{figure}

There are different ways of decomposing a rotation matrix in three Euler angles representations. In HPE the representation is built by rotating first around the $X$ axis of the camera coordinate system, pitch angle, then around the $Y$ axis, yaw angle, and then around the $Z$ axis, roll angle (see Fig.~\ref{fig:hpe_config}) and the head reference system is configured in such a way that the rotation matrix, $\mR$, of a head facing frontally to the camera is the identity. 

So, a rotation matrix is the product of three elementary rotations matrices that follow the pitch, yaw, roll order, producing

{\footnotesize
\[
    \mR ( \alpha_p, \alpha_y, \alpha_r) = 
    \begin{pmatrix}
        \cos{\alpha_r} & \sin\alpha_{r} & 0\\
        -\sin{\alpha_r} & \cos{\alpha_r} & 0 \\
        0 & 0 & 1
    \end{pmatrix}
    \begin{pmatrix}
        \cos{\alpha_y} & 0 & -\sin{\alpha_y}\\
        0 & 1 & 0 \\
        \sin{\alpha_y} & 0 & \cos{\alpha_y}
    \end{pmatrix}
    \begin{pmatrix}
        1 & 0 & 0\\
        0 & \cos{\alpha_p} & \sin{\alpha_p} \\
        0 & -\sin{\alpha_p} & \cos{\alpha_p}
    \end{pmatrix},
\]}
where $\alpha_y$, $\alpha_p$ and $\alpha_r$ refer to yaw, pitch and roll angles respectively.

The use of Euler angles in a WRHP setting introduces two problems: the gimbal lock and the lack of continuity in the representation.

The first issue occurs when the yaw angle reaches $\pm 90^o$. This causes the other two axes to align and the representation collapses because one degree of freedom is lost. Algebraically we get
\begin{align}
    \mR \left( p, \frac{\pi}{2},r \right) = 
    \begin{pmatrix}
        0 & \sin{(\alpha_p+\alpha_r)} & -\cos{(\alpha_p+\alpha_r)}\\
        0 & \cos{(\alpha_p+\alpha_r)} & \sin{(\alpha_p+\alpha_r)}\\
        1 & 0 & 0
    \end{pmatrix}.
    \label{eq:gimbal}
\end{align}
As we can see in Eq.~\eqref{eq:gimbal}, pitch and roll angles become indistinguishable and a given orientation may have infinite representations in terms of Euler angles that satisfy the linear relation
\begin{equation}
    \alpha_p + \alpha_r = \alpha.
    \label{eq:linear_relation}
\end{equation}

So, given any pitch and roll configuration satisfying Eq.~\eqref{eq:linear_relation}, we can immediately recover $\mR$ from Eq.~\eqref{eq:gimbal}. The gimbal lock is not a significant problem with Euler angles representation,
if our model is able to learn the dependency between the angles of the two aligned axes. Previous works~\cite{Hsu19, Cao21} changed the representation in SRHP because of gimbal lock. In Sect.~\ref{sec:experiments_synthetic} we show experimentally that a CNN using Euler angles representation is able to produce accurate estimations in a gimbal lock configuration. 

\begin{figure}
    \centering
    \includegraphics[width=0.8\textwidth]{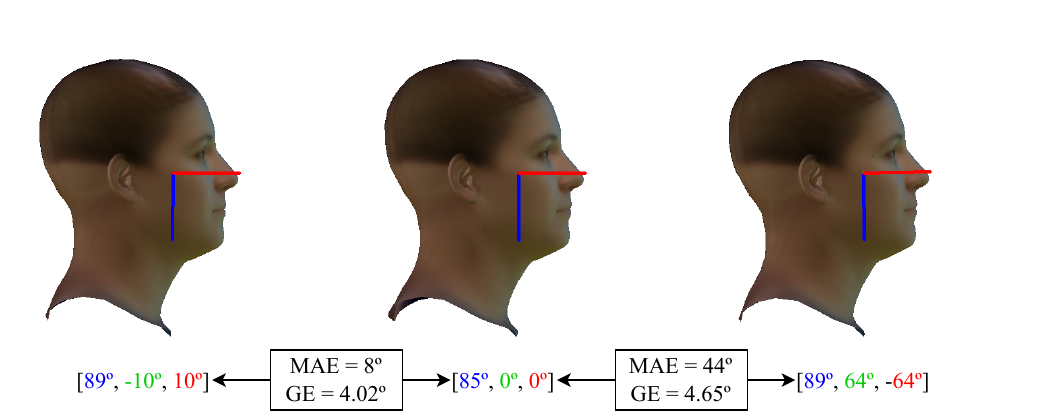}
    \caption{All faces have visually very similar configuration but MAE is very large due to gimbal lock. However, the geodesic distance is coherent. Color code: [{\color{blue} yaw}, {\color{green2} pitch}, {\color{red} roll}].}
    \label{fig:gimbal_lock_example}
\end{figure}

However, Euler angles cannot be used to measure the distance between two rotations. First because in the gimbal lock configuration one pose may be represented by an infinite number of different pitch and roll values. Second, in configurations close to the gimbal lock, \ie $\alpha_y=\frac{\pi}{2}+\delta$, for a small $\delta$,
\begin{align}
    \mR\left( p, \frac{\pi}{2}+\delta, r \right) = 
    \begin{pmatrix}
        -\delta\cos{\alpha_r} & \sin{(\alpha_p+\alpha_r)} & -\cos{(\alpha_p+\alpha_r)}\\
        \delta\sin{\alpha_r} & \cos{(\alpha_p+\alpha_r)} & \sin{(\alpha_p+\alpha_r)}\\
        1 & -\delta\sin{\alpha_p} & \delta\cos{\alpha_p}
    \end{pmatrix},
    \label{eq:gimbal_lock}
\end{align}
two close poses may be represented by very different Euler angles (see Fig.~\ref{fig:gimbal_lock_example}). This completely invalidates the use of any distance measure using Euler angles, such as $g_{MAE}$, $g_{MSE}$, $g_{EUC}$ and wrapped yaw distance, $g_w$, to compare two orientations either in SRHP or WRHP configurations. 
Conversely, the GE in Eq.~\eqref{eq:geodesic_error}, provides a coherent metric (see Fig.~\ref{fig:gimbal_lock_example}). 
Thus, the direct comparison of Euler angles should be abandoned in any HPE setting in favour of other well behaved metrics such as the geodesic distance.

\subsection{Discontinuity}

\begin{figure}
    \centering
    \includegraphics[width=0.65\textwidth]{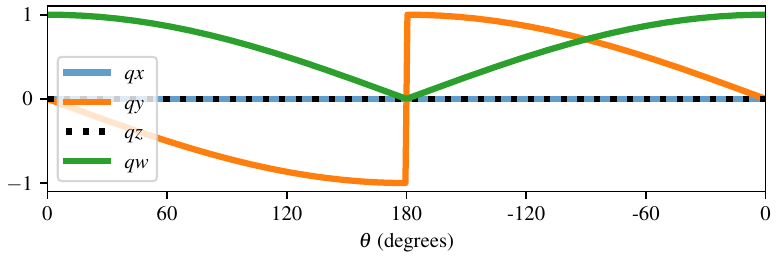}
    \caption{Discontinuity in quaternions under a rotation around the yaw axis (from $0^\circ$ to $360^\circ$). Component $q_y$ shows an abrupt change from -1 to +1 when the yaw reaches $180^\circ$.}
    \label{fig:quaternion_discontinuity}
\end{figure}

The gimbal lock prompted some HPE researchers to use alternative representations without that problem, such as quaternions~\cite{Hsu19}.
Both Euler angles and quaternions present discontinuity in their representation~\cite{Zhou19}. In Euler angles it appears when any angle reaches $\pm180^\circ$. In HPE this happens when the head rotates in the yaw angle half a circle. One component of quaternions also shows a discontinuity in the most extreme yaw configuration (see Fig.~\ref{fig:quaternion_discontinuity}).

The existence of a discontinuity affects learning because, near the discontinuity, similar facial appearances may have yaw angles (or quaternion components) with very different magnitude, therefore making learning more difficult~\cite{Zhou19}. 
Thus, discontinuity is a problem in a WRHP configuration but not in SRHP since, in the latter, faces are far from that problematic pose.

In Sect.~\ref{sec:experiments_synthetic} we show experimentally that for SRHP, Euler angles, quaternions and 6D representations achieve similar results. However, in a WRHP configuration, the continuous 6D representation provides significantly better results.

\subsection{Reference systems alignment for cross-data set evaluation}
\label{sec:nmethod_alignment}

\begin{figure}[htbp!]
    \centering
    \includegraphics[width=0.7\textwidth]{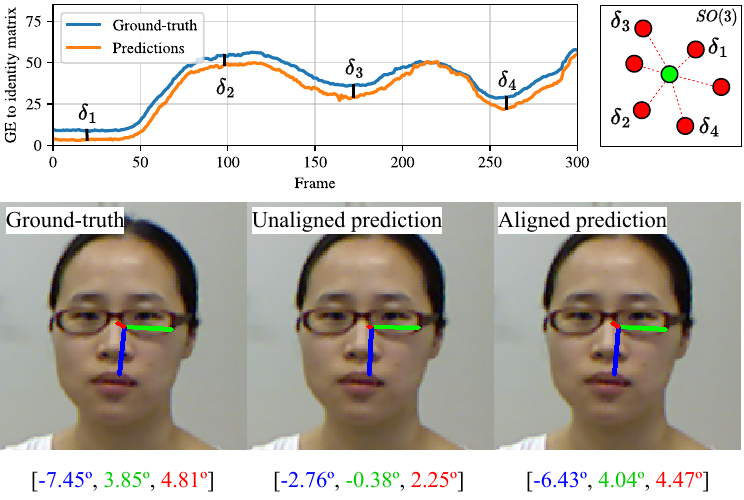}
    \caption{Reference systems alignment problem. Top: (left) Unaligned video sequence. Time and GE from identity shown in horizontal and vertical axes respectively; (right) Different alignment errors around the fixed alignment matrix. Bottom: Ground-truth, unaligned and aligned predictions in a specific frame with their corresponding Euler angles.}
    \label{fig:alignment}
\end{figure}

Data sets present built-in biases that negatively affect the accuracy of the usual intra-data set evaluation protocols~\cite{Torralba11}. This is especially pronounced when evaluating a HPE algorithm, and it is mostly caused by the biases produced by the annotation procedure~\cite{Valle21}. So, the most realistic approach to evaluate HPE involves a cross-data set approach, training the model in one data set and evaluating it in a different one~\cite{Valle21,Albiero21,Martyniuk22,Li23a,Hempel22}. This poses the additional problem of aligning the train and test coordinate reference systems. In Fig.~\ref{fig:alignment} we show this problem for a video sequence from the Biwi data set. In the plot we can see that the angular distances between the ground truth and the predictions are composed of a fixed gap, constant for all the sequence, and a random estimation noise. The fixed term is caused by the misalignment between the train and test data set reference systems. The random estimation noise is the HPE error we want to measure. Although cross-data set evaluation is a common practice in the community, this issue has been ignored in the literature. Here we propose a procedure to compensate this misalignment.

Let $\cR = \left\{ \mR_1, \dots, \mR_N \right\}$ be the set of ground truth rotation matrices of a data set with $N$ images, or video sequence with $N$ frames, and $\hat{\cR} = \left\{\hat{\mR}_1, \dots, \hat{\mR}_N \right\}$ be the set of predictions. If the estimations were perfect and the reference systems of train and test data sets were aligned, then $\mR_i^\top\hat{\mR}_i=\mI\,\forall i$, where $\mI$ is the identity matrix. However, $\hat{\mR}_i$'s are affected by an estimation error matrix different for each image, $\ttdelta\mR_i$, and an alignment matrix, $\ttDelta$, common to all estimations in the set ${\hat{\cR}}$, so $\hat{\mR}_i\,\ttdelta\mR_i\,\ttDelta = \mR_i\,\forall i$. To estimate $\ttDelta$ we assume the $\ttdelta\mR_i$'s are random perturbations around the identity, $\mI$, caused by a combination of different HPE errors. Hence, $\ttdelta_i=\ttdelta\mR_i\,\ttDelta = \hat{\mR}_i^\top\mR_i$ are randomly distributed around $\ttDelta$ (see Fig.~\ref{fig:alignment} top right). So, we can estimate the alignment matrix $\ttDelta$ as the mean of all $\ttdelta_i$'s. This is a \emph{single rotation averaging} problem that can be computed with the Karcher mean~\cite{Hartley13}
\[
\hat\ttDelta = \argmin_{\mR\in SO(3)} \sum_{i=1}^N f_{GE} \left( \mR, \ttdelta_i \right)^2,
\]
for which there is a simple and convergent algorithm~\cite{Manton04}.

Finally, we compute a new set of aligned predictions $\hat{\cR}_a = \left\{\hat{\mR}_1\hat\ttDelta^\top, \dots, \hat{\mR}_N\hat\ttDelta^\top \right\}$
that we use to evaluate the HPE.


\section{Opal loss function for HPE}
\label{sec:method_opal}

In this section we introduce a new OPtimAL loss function, Opal, based on a generalization of the geodesic loss in Eq.~\eqref{eq:geodesic_error} to improve the performance of HPE. It uses the geodesic error of each training sample to control its  contribution to the learning, optimizing for example the accuracy for a specific subset of images.

\begin{figure}
    \centering
    \includegraphics[width=0.8\textwidth]{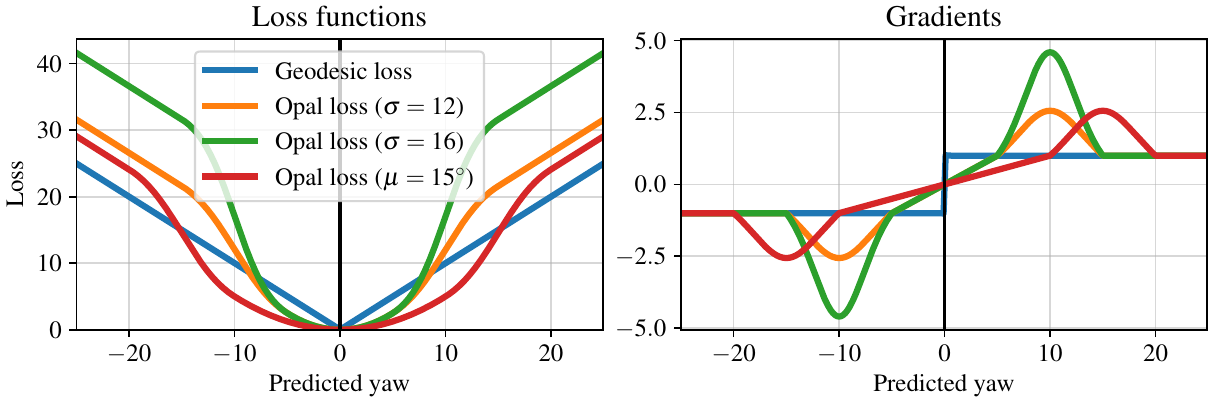}
    \caption{Comparison of Geodesic and Opal losses, and the influence functions (gradients) obtained by both. The horizontal axis represent the predicted yaw (pitch and roll are fixed to 0$^{\circ}$), considering a ground-truth of $\left[ 0^\circ, 0^\circ, 0^\circ \right]$.}
    \label{fig:loss_plot}
\end{figure}

The gradient of the distance used in the loss function, typically denoted as the influence function, drives the minimisation process and determines the importance of each sample. As shown in Fig.~\ref{fig:loss_plot} (left), the geodesic loss resembles an L1. Inspired by~\cite{Feng18a} we design Opal in such way that its influence function (Fig.~\ref{fig:loss_plot} (right)) has a pseudo-Gaussian distribution that lets us specify which range of errors are more important, and by how much, while also linearly reducing the influence of small errors. 
Furthermore, we set the influence of large errors to 1, to avoid exploding gradients early in the training phase.

The Opal generalization to the geodesic distance between two rotation matrices is piece-wise function,
\begin{align}
    g_{opal}(\hat{\mR}, \mR) = \left\{
    \begin{array}{ll}
        a \cdot G^2 + b & \quad G < \epsilon \\
        c \cdot (\tanh{(\sigma \cdot G - \mu)} + tanh(\mu)) & \quad \epsilon \geq G < \beta \\
        G + d & \quad G \geq \beta,
    \end{array}
    \right.
\end{align}
where $G = g_{GE}(\hat{\mR}, \mR)$; $\epsilon$ and $\beta$ represent the angular thresholds between L2 and $tanh$, and $tanh$ and L1 functions, respectively; $\mu$ and $\sigma$ control the mean and height of the pseudo-Gaussian
influence function; and $a$, $b$, $c$ and $d$ are constants that ensure continuity and differentiability. Finally, the Opal loss is defined by extending $g_{opal}$ to the training set,
\begin{equation}
    f_{opal} = \frac{1}{N} \sum_{i=1}^N 
     g_{opal}(\hat{\mR}_i, \mR_i).
\end{equation}

As we can see in Figures~\ref{fig:loss_plot} and~\ref{fig:opal_gradients}, the parameters of Opal loss can be adjusted to fit the distribution of the Geodesic Errors in the data set. This improves the generalization power of the network by balancing the contribution of different samples given their difficulty in training. In Section~\ref{sec:experiments_wide_hpe} we empirically demonstrate the usefulness of this loss function for wide-range HPE.

\section{Experiments}
\label{sec:experiments}

In this section we test different combinations of head pose representations and loss functions. We use three configurations for HPE: 1) Euler angles and RMSE loss, defined as $g_{RMSE}(\hat\vp, \vp)=\sqrt{g_{MSE}(\hat\vp, \vp)}$,
denoted as \emph{Euler} in our experiments, 2) quaternions with RMSE loss,
denoted as \emph{Quaternion}, and 3) 6D with geodesic error loss (Eq.~\ref{eq:geodesic_error})
denoted simply as \emph{6D}.

\subsection{Data sets}
\label{sec:experiments_datasets}

300W-LP and AFLW2000-3D are SRHP data sets introduced in~\cite{Zhu19b} consisting of 61225 and 2000 images acquired from 300W~\cite{Sagonas16} and AFLW~\cite{Kostinger11}, respectively. They include automatically annotated 3D landmarks and head poses generated by fitting a 3DMM to each sample image. We follow the standard protocol in AFLW2000-3D~\cite{Ruiz18}, and discard 30 images with yaw, pitch or roll outside the range [-99$^{\circ}$, 99$^{\circ}$]. 

Biwi~\cite{Fanelli13} is also a SRHP data set that contains 15677 frames from 24 videos of 20 subjects. It includes head pose annotations and facial depth masks obtained by a Kinect device without face bounding box annotations. In our experimentation we also follow common protocols presented in~\cite{Yang19} using 300W-LP as train set and 13219 frames of Biwi as test set, with bounding boxes provided by the MTCNN face detector~\cite{Zhang16b}. 

CMU Panoptic~\cite{Joo17} is a WRHP data set acquired in laboratory conditions that contains $84$ video sequences organized into $8$ groups, each one focusing on different scenarios, such as \textit{range of motion}, \textit{musical instruments}, \textit{social games}, \textit{haggling}, \textit{dance}, \textit{toddler} and \textit{others}. Each video sequence consists of $31$ HD quality videos with a resolution of $1920\times 1080$ pixels taken from a set of calibrated cameras located in a dome structure covering a full-hemisphere.
Unfortunately, Panoptic does not include pose and bounding box annotations. Zhou \etal~\cite{Zhou20} used the annotated 3D landmarks to generate them, but the training/testing protocols were not released. For this reason, we create a public benchmark containing $12$ video sequences from \textit{range of motion} and $5$ video sequences from \textit{haggling}. The number of images for training, validation and testing are, respectively, 61008, 22940 and 25141. 
We discarded the frames where the landmarks could not be correctly projected onto all cameras.
In the \textit{haggling} sequences, where more than one subject is present in the same scene, we only considered the first person appearing in the annotation files. 

\subsection{Implementation details}
\label{sec:experiments_implementation}

During training, we perform data augmentation by applying the following random operations: horizontal flip, in plane rotations~\cite{Sheka21}, scaling, synthetic occlusions, HSV color space manipulation and image blurring. 

We build an encoder-like neural network which extracts relevant features from a 128$\times$128 RGB input image. The full architecture consists of 6 layers of stride 2 convolutions. At each one, we introduce ``inverted residual bottleneck'' modules from MobileNet-V2~\cite{Sandler18}. As usual, each conv layer has batch normalization and a ReLU6 activation function. This is followed by a global average pooling layer to finally obtain a 1$\times$1$\times$256 tensor which is fed into a last fully connected layer responsible of estimating the $P$ pose parameters. 

Our networks are implemented in Pytorch and use Adam optimizer with an initial learning rate set to $10^{-4}$. We always select the model parameters with lowest validation error. For 300W-LP, we fine-tune the weights of a model pre-trained in the landmark detection task similar to~\cite{Valle21}. We train the Euler model for 600 epochs, dropping the learning rate to $10^{-5}$ at epoch 300, and the 6D model for 900 epochs dropping the learning rate from $5 \cdot 10^{-4}$ to $5 \cdot 10^{-5}$ at epoch 450. For Panoptic, the learning rate is halved after 30 epochs without improvement in the validation. We stop training when the validation error stops improving for 60 epochs.

\subsection{Synthetic experiments} 
\label{sec:experiments_synthetic}

\begin{figure}
    \centering
    \includegraphics[width=0.8\textwidth]{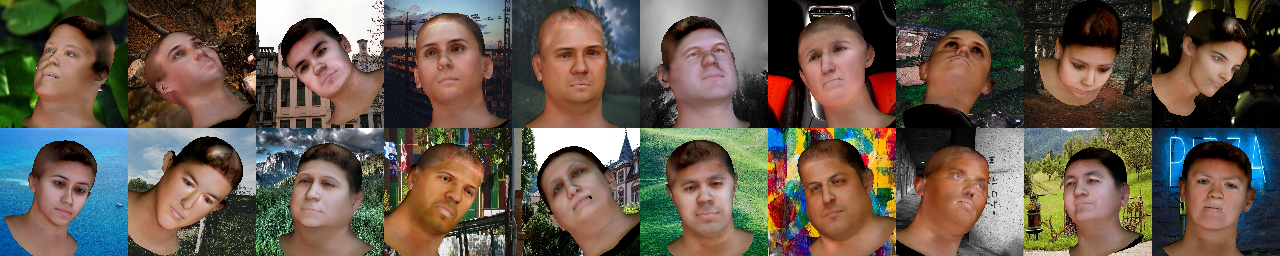}
    \caption{Representative examples of the synthetic face images generated.}
    \label{fig:flame_samples}
\end{figure}

In this section, we create a synthetic WRHP estimation benchmark to compare Euler, Quaternion and 6D representations. We use the Flame 3D model~\cite{Li17} with the albedo subspace of the Basel Face Model~\cite{Paysan09} to generate synthetic faces with head pose ranging between six intervals of yaw angles, \ie [$-30^{\circ}$,$30^{\circ}$], [$-60^{\circ}$,$60^{\circ}$], [$-90^{\circ}$,$90^{\circ}$], [$-120^{\circ}$,$120^{\circ}$], [$-150^{\circ}$,$150^{\circ}$] and [$-180^{\circ}$,$180^{\circ}$], while pitch and roll angles have a fixed range of [$-45^{\circ}$,$45^{\circ}$], and neutral expressions. We add more realism to the generated images, using backgrounds taken from the BG-20K data set~\cite{Li22a}. For each yaw interval, its corresponding data set contains 128000 training images, 12800 validation images and 38400 test images.
In Fig.~\ref{fig:flame_samples} we display some representative examples of the synthetic face images generated. 

\begin{figure}
    \centering
    \includegraphics[width=0.8\textwidth]{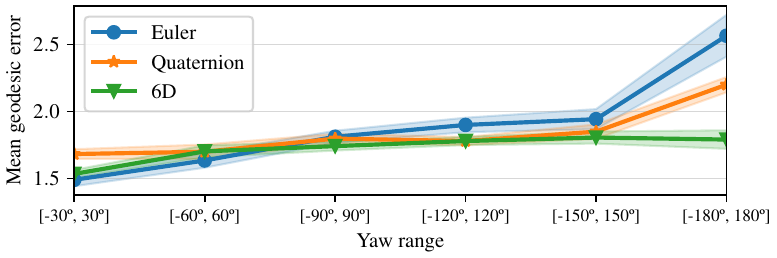}
    \caption{Head pose representation experiment. %
    Blue, orange and green colours compare the GE obtained at each interval using Euler, quaternion and 6D representations respectively. We also display with shading three standard deviations of the error.}
    \label{fig:flame_results}
\end{figure}

In the first experiment (see Fig.~\ref{fig:flame_results}), we train 3 different models for each interval. To make a fair comparison among the three representations, we use the GE as evaluation metric and the same loss function on all networks, $g_{RMSE}$.
As a global trend, we can see that the mean head pose errors increase as the yaw interval widens. This was expected since a wider yaw range requires the model to generalize across a broader spectrum of pose variations, making the learning task more challenging. 

The gimbal lock has been traditionally perceived as a limitation of Euler angles for HPE~\cite{Hsu19,Abate22,Hempel22,Zhou20,Cao21}. Indeed, as discussed in Sect.~\ref{sec:method}, Euler angles cannot be used to measure the distance between two rotations, but in certain settings they are a good representation choice.
Our results in Fig.~\ref{fig:flame_results} demonstrate that in a SRHP configuration, with yaw in the range $[-90^\circ,90^\circ]$, the performance difference among all three competing representations is marginal. To confirm this result we perform a second experiment in which we decouple the inherent difficulty of predicting pose in profile faces from the gimbal lock. To this end we train our model to estimate the pose of semi-frontal faces with yaw angles ranging between [$-10^{\circ}$,$10^{\circ}$], but re-annotated as [$80^{\circ}$,$100^{\circ}$], so that all labels are close to the gimbal lock. In Table~\ref{tab:flame_gimbal_lock} we compare the results of this experiment for Euler angles and 6D representations. In terms of the GE error, both models have a very similar performance, with a marginal edge towards Euler. These results prove experimentally what we had discussed in Sect.~\ref{sec:method}, namely, in SRHP configuration Euler angles are a good representation, if the model is able to learn linear dependency between pitch and roll angles in the gimbal lock.
\begin{table}
    \centering
    \footnotesize
    \begin{tabular}{l|ccc|c|c}
        \toprule
        \multirow{2}{*}{Parameterization} & \multicolumn{4}{c|}{MAE} & \multirow{2}{*}{GE} \\
        {} & Yaw & Pitch & Roll & Mean \\
        \midrule
        Euler & \first{1.07} & \first{1.65} & \first{0.60} & \first{1.10} & \first{2.24}  \\
        \midrule
        6D & 1.12 & 3.93 & 3.36 & 2.80 & 2.32 \\
        \bottomrule
    \end{tabular}
    \caption{MAE and GE in our synthetic data set with frontal faces and gimbal lock.}
    \label{tab:flame_gimbal_lock}
\end{table}

Differently to the gimbal lock case, we can also see in Fig.~\ref{fig:flame_results} that the prediction error for the Euler and quaternions models grow very rapidly when the range of yaw rotations is $[-180^\circ,180^\circ]$. The discontinuity of Euler and quaternions representations near the yaw $\pm180^\circ$ limits their performance. So, the 6D representation, since it is continuous, provides the best performance in the in the WRHP setting. 

\subsection{SRHP results} 
\label{sec:experiments_short_hpe}

In this section we perform experiments on a SRHP configuration to confront the different representations with the existing literature. The most realistic SRHP evaluation is based on a cross-data set methodology. A widely used benchmark uses 300W-LP as training set and AFLW2000-3D and Biwi as test sets. To train the 300W-LP model, we shuffle the training set and split it into 90\% train and 10\% validation subset. We also crop faces using the bounding box of the landmarks annotations enlarged by 60\%. 

In Table~\ref{tab:sota_comparison_aflw2000} we compare our model using two representations with the SOTA in AFLW2000-3D. It is difficult to extract a conclusion about different representations since some methods use 3DMMs, \eg DAD-3DNet~\cite{Martyniuk22}, DSFNet~\cite{Li23a}, others additional training data,\eg img2pose~\cite{Albiero21}, DAD-3DNet~\cite{Martyniuk22}, and, as expected, there are different CNN architectures. %
To address this issue we evaluate our model using the same network architecture, training images and data augmentation, only changing the representation. Our unaligned results in Table~\ref{tab:sota_comparison_aflw2000}
confirm the synthetic experiments. Euler and 6D representations provide the same results using the GE metric, with a marginal edge in favor of the former. To compare with the literature we also provide the MAE.

\begin{table}
    \footnotesize
    \centering
    \begin{tabular}{l|c|ccc|c|c}
        \toprule
        \multirow{2}{*}{Method} & \multirow{2}{*}{Representation} & \multicolumn{4}{c|}{MAE ($\downarrow$)} & \multirow{2}{*}{GE ($\downarrow$)}\\
         & & yaw & pitch & roll & mean & \\
        \midrule
        HopeNet~\cite{Ruiz18} & Euler & 6.47 & 6.56 & 5.44 & 6.15 & 9.93\\
        FSA-Net~\cite{Yang19} & Euler & 4.50 & 6.08 & 4.64 & 5.07 & 8.16\\
        WHENet~\cite{Zhou20} & Euler & 4.44 & 5.75 & 4.31 & 4.83 & -\\
        TriNet~\cite{Cao21} & Rot. matrix & 4.19 & 5.76 & 4.04 & 4.66 & -\\
        TokenHPE~\cite{Zhang23} & Rot. matrix & 4.36 & 5.54 & 4.08 & 4.66 & -\\
        QuatNet~\cite{Hsu19} & Quaternion & 3.97 & 5.61 & 3.92 & 4.50 & -\\
        MFDNet~\cite{Liu21a} & Rot. matrix & 4.30 & 5.16 & 3.69 & 4.38 & -\\
        img2pose~\cite{Albiero21} & Rot. vector & 3.42 & 5.03 & 3.27 & 3.91 & \third{6.41}\\
        MNN~\cite{Valle21} & Euler & 3.34 & 4.69 & 3.48 & 3.83 & -\\
        DAD-3DNet~\cite{Martyniuk22} & 6D & 3.08 & 4.76 & 3.15 & 3.66 & -\\
        DSFNet~\cite{Li23a} & Rot. matrix & \first{2.65} & \second{4.28} & \second{2.82} & \first{3.25} & -\\ 
        \textbf{Ours (unaligned)} & Euler & \second{2.76} & \first{4.25} & \first{2.76} & \second{3.26} & \first{5.29}\\
        \textbf{Ours (unaligned)} & 6D & 2.85 & \third{4.59} & \third{3.04} & \third{3.49} & \second{5.37}\\
        \midrule
        \textbf{Ours (aligned)} & Euler & 2.75 & 4.23 & 2.76 & 3.25 & 5.28\\
        \textbf{Ours (aligned)} & 6D & 2.83 & 4.55 & 3.04 & 3.47 & 5.34\\
        \bottomrule
    \end{tabular}
    \caption{SRHP results using AFLW2000-3D. GE is only available in methods that provide test code. Results ranked \first{first}, \second{second} and \third{third} are shown respectively in blue, green and red colors.}
    \label{tab:sota_comparison_aflw2000}
\end{table}

The alignment procedure introduced Sect.~\ref{sec:nmethod_alignment} does not provide a reduction of the GE error in this experiment. This is an expected result since the training, 300W-LP, and test data sets, AFLW2000-3D, use the same annotation algorithm
and, hence, their reference systems are aligned. Moreover, these results prove that our alignment procedure does not influence the result, if the reference systems of the training and test data sets are aligned.

\begin{table}
    \centering
    \footnotesize
    \begin{tabular}{l|c|ccc|c|c}
        \toprule
        \multirow{2}{*}{Method} & \multirow{2}{*}{Representation} & \multicolumn{4}{c|}{MAE ($\downarrow$)} & \multirow{2}{*}{GE ($\downarrow$)}\\
         & & yaw & pitch & roll & mean & \\
        \midrule
        HopeNet*~\cite{Ruiz18} & Euler & 4.81 & 6.61 & 3.27 & 4.89 & 9.53\\
        QuatNet*~\cite{Hsu19} & Quaternion & 4.01 & 5.49 & 2.93 & 4.14 & -\\
        FSA-Net~\cite{Yang19} & Euler & 4.27 & 4.96 & 2.76 & 4.00 & 7.64\\
        DAD-3DNet~\cite{Martyniuk22} & 6D & 3.79 & 5.24 & 2.92 & 3.98 & -\\
        TriNet~\cite{Cao21} & Rot. matrix & 4.11 & 4.75 & 3.04 & 3.97 & -\\
        img2pose~\cite{Albiero21} & Rot. vector & 4.56 & \first{3.54} & 3.24 & 3.78 & \first{7.10}\\
        TokenHPE~\cite{Zhang23} & Rot. matrix & 3.95 & \third{4.51} & \third{2.71} & 3.72 & -\\
        MNN*~\cite{Valle21} & Euler & 3.98 & \third{4.61} & \first{2.39} & \third{3.66} & -\\
        MFDNet~\cite{Liu21a} & Rot. matrix & \first{3.40} & 4.68 & 2.77 & \second{3.62} & -\\
        WHENet*~\cite{Zhou20} & Euler & \third{3.60} & \second{4.10} & \third{2.73} & \first{3.48} & -\\
        \textbf{Ours (unaligned)} & Euler & 4.54 & 5.05 & 2.80 & 4.13 & \third{7.49}\\
        \textbf{Ours (unaligned)} & 6D & 4.58 & 4.65 & \third{2.71} & 3.98 & \second{7.30}\\
        \midrule
        HopeNet*~\cite{Ruiz18} (aligned) & Euler & 4.53 & \second{3.08} & 2.83 & 3.48 & 6.60\\
        FSA-Net~\cite{Yang19} (aligned) & Euler & \first{3.59} & \first{2.90} & \first{2.27} & \first{2.92} & \first{5.36}\\
        img2pose~\cite{Albiero21} (aligned) & Rot. vector & \third{4.04} & 3.12 & 3.03 & 3.40 & 6.23\\
        \textbf{Ours (aligned)} & Euler & \second{3.98} & \third{3.09} & \second{2.40} & \second{3.16} & \second{5.42}\\
        \textbf{Ours (aligned)} & 6D & \second{3.98} & 3.24 & \third{2.42} & \third{3.21} & \third{5.48}\\
        \bottomrule
    \end{tabular}
    \caption{SRHP results using Biwi. GE only available in methods that provide test code. * means methods where MTCNN detections are not used or is not stated. Results ranked \first{first}, \second{second} and \third{third} are shown respectively in blue, green and red colors.}
    \label{tab:sota_comparison_biwi}
\end{table}

In Table~\ref{tab:sota_comparison_biwi} we compare with the SOTA in Biwi. 
Like in the previous experiment, the results of models using Euler angles and 6D representation are very close. However, in this case and in contradiction with previous experiments, our model with 6D representation seems to be slightly better than Euler. The problem here, and with all previous results in the literature, is that we are ignoring the misalignment between the train and test data sets reference systems. Biwi was annotated by fitting a 3D model to the point cloud of a kinect depth map via ICP, whereas 300W-LP was labeled with the 3DDFA CNN~\cite{Zhu19b}.
In this case, the use of the alignment procedure reduces in 25\% the GE error of our model using a 6D representation. Moreover, in the aligned results our Euler model is again marginally better, which is in agreement with all previous experiments.

In Table~\ref{tab:sota_comparison_biwi} we also show aligned results of previous methods. In all of them we provide a more accurate estimation that significantly reduces the Geodesic Error (GE) and, what is more interesting, the ranking of algorithms also changes. 

In this section we confirmed experimentally some of the methodological results presented in Sect.~\ref{sec:method}. In a SRHP configuration Euler angles and 6D representations have similar performance.
However, for cross-data set evaluation, it is required to apply an alignment procedure to remove systematic errors.

\subsection{WRHP results} 
\label{sec:experiments_wide_hpe}

In this section we perform experiments on a WRHP configuration to confirm the results obtained using synthetic images in Sect.~\ref{sec:experiments_synthetic}. Here we also experiment with our novel Opal loss that allows us to emphasise certain error ranges while training.

The first challenge with WRHP experiments is the availability of annotated real images. Zhou~\etal~\cite{Zhou20} proposes a WRHP protocol using a combined data set of 300W-LP and Panoptic. 
However, 
annotations are not public. 
We propose a new benchmark based exclusively on Panoptic images (see Sect.~\ref{sec:experiments_datasets}).

\begin{figure}
    \centering
    \includegraphics[width=0.70\textwidth]{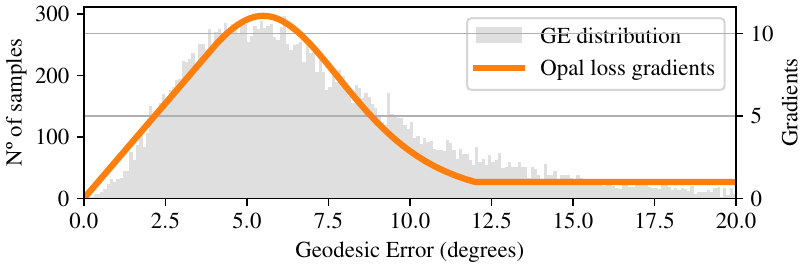}
    \caption{Gradients of Opal loss (in orange) plotted over the GE distribution of the validation set of CMU Panoptic using a 6D model with geodesic loss (in gray).}
    \label{fig:opal_gradients}
\end{figure}

In our experiments we train with Euler, quaternion and 6D representations. Additionally, we use Opal to improve the GE of the 6D network. Since what we estimate in Eq. \ref{eq:geodesic_error} is the mean error of all training samples, we set the parameters of Opal so that its influence function fits the distribution of errors of the validation set (see Fig. \ref{fig:opal_gradients}). In this way our model concentrates 
more in the GE range around 5.5, and does not waste its learning capacity trying to improve very easy or atypical cases, below 2 or above 12 degrees respectively. 

Results are shown in Table~\ref{tab:panoptic-results}, aggregated (Mean column) and broken down into frontal ($\lvert \alpha_y \rvert \in [0^\circ, 60^\circ]$), profile ($\lvert \alpha_y \rvert \in [60^\circ, 120^\circ]$) and back views ($\lvert \alpha_y \rvert \in [120^\circ, 180^\circ]$).
Note that, since WHENet does not provide training code, we used the public pre-trained weights to evaluate their network. In contrast, we trained 6DRepNet with their training code, choosing the epoch with minimum validation loss.
The first observation in this experiment is that here the mean GE is greater than in the SRHP case shown in Tables~\ref{tab:sota_comparison_aflw2000} and~\ref{tab:sota_comparison_biwi}, which means that this problem is more difficult. 
As expected from the synthetic experiments, and due to the discontinuity effect, 
the Mean error of Euler and quaternion representations is respectively 35\% and 21\% larger than that of the 6D representation. Here again, like in the synthetic tests, Euler provides the worst results.
6DRepNet results show a similar effect, since they are close to our 6D network, and also outperform our Euler and Quaternion networks. We can conclude that
a key factor in WRHP estimation is choosing a continuous representation.

\begin{table}
    \centering
    \footnotesize
    \begin{tabular}{l||c|c|c|c|c}
        \toprule
        Method & Representation & Mean & Frontal & Profile & Back \\
        \midrule
        WHENet~\cite{Zhou20} & Euler & 24.83 & 29.34 & 24.73 & 20.33 \\ 
        6DRepNet~\cite{Hempel22} & 6D & \third{8.08} & \second{5.80} & \third{8.07} & \third{10.40} \\
        \hline
        \textbf{Ours} & Euler & 10.47 & 7.69 & 10.08 & 13.68 \\
        \textbf{Ours} & Quaternion & 9.32 & 7.04 & 8.98 & 11.98 \\
        \textbf{Ours} & 6D &  \second{7.70} & \third{5.81} &  \second{7.15} & \first{10.18} \\
        \textbf{Ours + Opal loss} & 6D & \first{7.45} & \first{5.40} & \first{6.75} &  \second{10.25} \\
        \bottomrule
    \end{tabular}
    \caption{
    WRHP results in 
    CMU Panoptic benchmark. 
    Mean GE and GE in frontal ($\lvert yaw \rvert \in \left[ 0^{\circ}, 60^{\circ} \right]$), profile ($\lvert yaw \rvert \in \left[ 60^{\circ}, 120^{\circ} \right]$) and back ($\lvert yaw \rvert \in \left[ 120^{\circ}, 180^{\circ} \right]$) views. \first{first}, \second{second} and \third{third} are shown respectively in blue, green and red colors.}
    \label{tab:panoptic-results}
\end{table}

\begin{figure}
    \centering
    \includegraphics[width=0.65\textwidth]{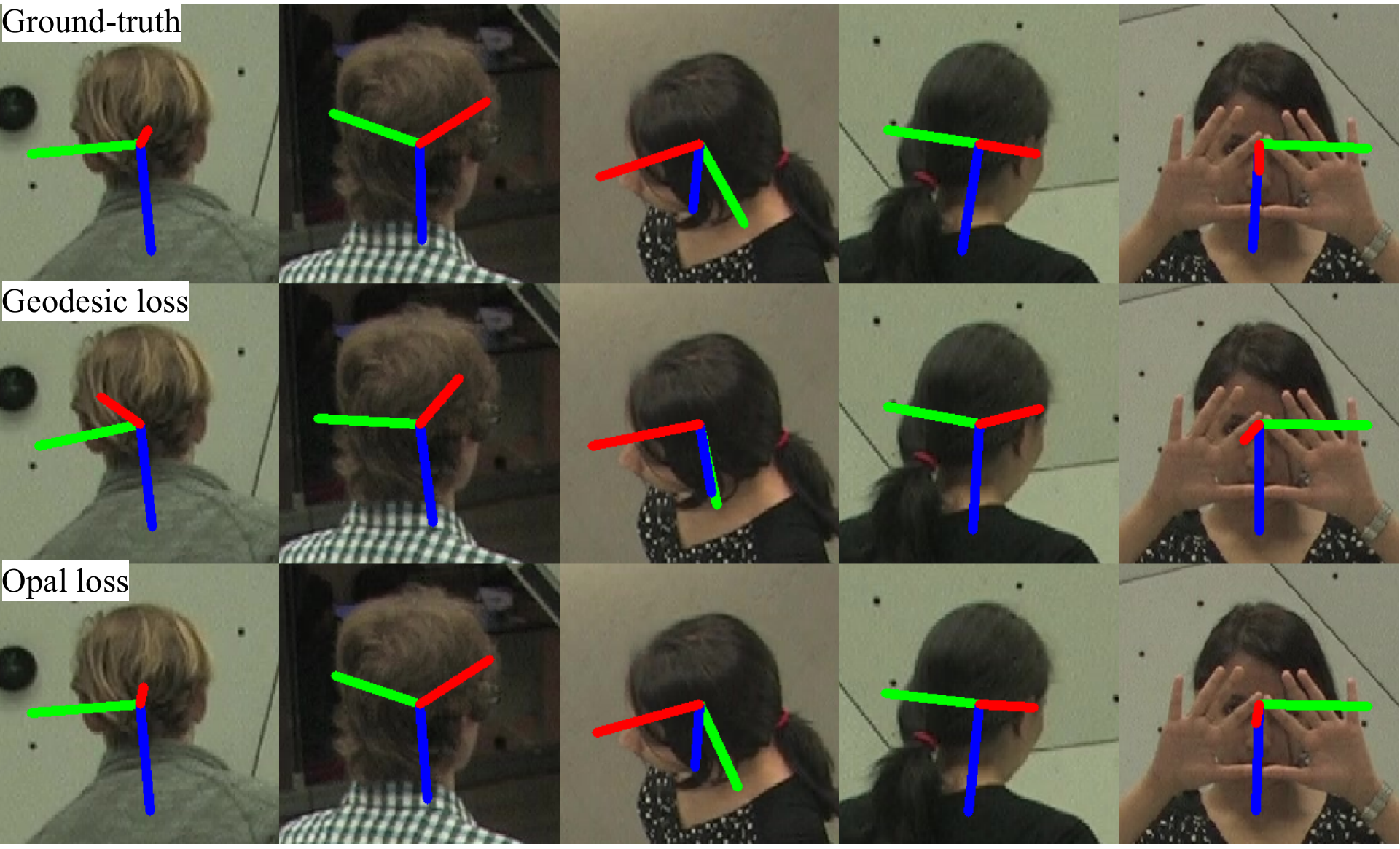}
    \caption{Qualitative comparison between annotations (first row) and predictions using Geodesic loss (second row) and Opal loss (third row) in CMU Panoptic test set.
    }
    \label{fig:panoptic_samples}
\end{figure}

Furthermore, Opal loss is capable of improving the mean GE of the 6D network by focusing more on the most frequent errors (mostly related to frontal and profile views) and decreasing the importance of small and atypical errors. In Fig.~\ref{fig:panoptic_samples} we show some examples for which Opal significantly improved the estimation.

\section{Conclusions}
\label{sec:conclusions}

This paper addressed the problem of estimating the orientation of a head in an image both in the profile to frontal SRHP and in the full 360º rotation WRHP setting. 
In our analysis we distinguish the representation used for the pose from the distance between two orientations used as error metric. This enabled us to shed some light on the adequacy of each representation and error metric.
Contrary to common believe, we show that for HPE the gimbal lock is not a drawback for Euler angles representation. In fact, given its minimal dimensionality, intuitive interpretation, and marginal edge over other alternatives, it is the preferred representation in a SRHP setting involving frontal and profile heads.
However, to solve the HPE in the full WRHP setting we need a continuous representation, like the 6D obtained with two columns of the rotation matrix~\cite{Zhou19}.

Unlike with the representation problem, the gimbal lock prevents the use of Euler angles for measuring the distance between two orientations. So, the popular MAE metric should be avoided both in SRHP and WRHP settings. We suggest the use of proper angular metric such as the geodesic error, GE~\cite{Huynh09,Hartley13}, because of its interpretability.

For a proper cross-data set evaluation we must consider the fact that it is very likely that the reference systems used in the annotations of the training and test data sets are different. 
As a consequence, we can not directly compare the orientation estimated by the trained model with the ground truth annotation in the test data set. We have also introduced a procedure to align the estimations with the ground truth annotations and empirically show that it does not bias the estimation in cases where the train and test annotations share the same reference system. With this procedure we set a new, more accurate, SOTA in the 300W-LP/Biwi cross-data set benchmark.

We also generalize the geodesic distance to introduce Opal, a new distance conceived for controlling the contribution of images in the train data set to the learning process. In the future we plan to use this distance in other regression problems, such as facial landmark estimation and to optimize other particular ranges of error, instead of the usual average.

Finally, we propose a new benchmark based on the Panoptic data set and the annotations proposed in WHENet~\cite{Zhou20}. 

\section*{Acknowledgements}

This work was partially funded by project PID2022-137581OB-I00 from MCIN/AEI/10.13039/501100011033/FEDER, UE. Alejandro Cobo was also funded by the Comunidad de Madrid grant PEJ-2020-AI/TIC-17682, and with a doctoral contract from Universidad Polit\'ecnica de Madrid, UPM. Jos\'e M. Buenaposada was partially funded by ``AYUDA PUENTE 2022'', Universidad Rey Juan Carlos (ref. M3037). Luis Baumela and Jos\'e M. Buenaposada are members of ELLIS Unit Madrid, funded by the Autonomous Community of Madrid. The authors gratefully acknowledge the UPM for providing computing resources on the Magerit supercomputer.       

\bibliographystyle{plain} 
\bibliography{faces}

\end{document}

%% file: pcr_defs.tex
\def\mat#1{\mathchoice{\mbox{\boldmath $\displaystyle\tt#1$}}
{\mbox{\boldmath$\textstyle\tt#1$}}
{\mbox{\boldmath$\scriptstyle\tt#1$}}
{\mbox{\boldmath$\scriptscriptstyle\tt#1$}}}

\def\vect#1{\mathchoice{\mbox{\boldmath $\displaystyle\bf#1$}}
{\mbox{\boldmath  $\textstyle\bf#1$}}
{\mbox{\boldmath  $\scriptstyle\bf#1$}}
{\mbox{\boldmath  $\scriptscriptstyle\bf#1$}}}

\def\vd{{\vect d}}

\def\vp{{\vect p}}

\def\vy{{\vect y}}

\def\v0{{\vect 0}}

\def\mDelta{{\mat\mDelta}}

\def\m1{{\mat 1}}

\def\mI{{\mat I}}

\def\mR{{\mat R}}

\def\cR{\mathcal{R}}

\def\argmin{\mbox{arg}\min}
